# REVIEW ON DEEP LEARNING TECHNIQUES FOR UNDERWATER OBJECT DETECTION

Radhwan Adnan Dakhil and Ali Retha Hasoon Khayeat

Department of Computer Science, University of Kerbala, Karbala, Iraq

## ABSTRACT

*Repair and maintenance of underwater structures as well as marine science rely heavily on the results of underwater object detection, which is a crucial part of the image processing workflow. Although many computer vision-based approaches have been presented, no one has yet developed a system that reliably and accurately detects and categorizes objects and animals found in the deep sea. This is largely due to obstacles that scatter and absorb light in an underwater setting. With the introduction of deep learning, scientists have been able to address a wide range of issues, including safeguarding the marine ecosystem, saving lives in an emergency, preventing underwater disasters, and detecting, spooring, and identifying underwater targets. However, the benefits and drawbacks of these deep learning systems remain unknown. Therefore, the purpose of this article is to provide an overview of the dataset that has been utilized in underwater object detection and to present a discussion of the advantages and disadvantages of the algorithms employed for this purpose.*

## 1. INTRODUCTION

Algorithms for accurately detecting and recognizing objects in images and real-world data are used for tasks such as tracking the location, motion, and orientation of objects. For an object to be detected and recognized, the algorithm must determine whether or not an object or objects are present. Object detection is "the process of accurately identifying an object, localizing that object inside an image, and performing semantic or instance segmentation [1]. The problem statement for object detection is to figure out where objects are in an image (called "object localization") and what class each point belongs to (object classification).Object classification, selection of an informative region, and feature extraction are the three main components that comprise the pipeline of traditional object detection models.

1) Selecting Informative Regions – Since objects can appear anywhere in the frame and in various sizes, it makes sense to deploy a sliding window with many scales to search the full image.
2) Extracting Features – Identifying a variety of objects requires the extraction of visual features that can provide a semantic and robust representation. Representative features include SIFT [2], HOG [3], and Haar-like [4]. As a result of their ability to produce representations associated with complex brain cells [2], these features are important.
3) Classification – In addition to differentiating a target object from all other categories, a classifier is required to make representations more hierarchical, semantic, and informative for visual identification. Among the many choices available, the Supported Vector Machine (SVM) [5], the AdaBoost [6], and the Deformable Part-based Model (DPM) [7] are commonly suggested.

Object recognition of computer vision is a method to determine the identity of an object was seen in still images or moving videos. It involves distinguishing between two targets that are extremely similar as well as between one, two, or even more types of targets that are depicted in an image. Object recognition's ultimate objectives are to first recognize objects within an image in the same way that humans do and then to train a computer to acquire some level of image comprehension. The same object can be recognized when viewed from a variety of perspectives, including front, rear and side views. Additionally, the object can be identified whether it is a different size or when there is some obstruction between the viewer and the object [8]. In recent years, numerous object recognition tasks, such as handwriting [9, 10], license plate recognition [11], speech recognition [12], lane line recognition [13], face recognition [14], ship and military object recognition [15, 16], fish and underwater creature recognition [17, 18], etc., have been the subject of extensive research. Even though the oceans occupy approximately two-thirds of the globe, relatively few technologies related to marine research have been investigated to a sufficient degree [19, 20]. Feature extraction and classification are the two essential phases that comprise marine object recognition from a practical standpoint. Nevertheless, feature extraction is the more important of the two steps. The processes of pre-processing, feature extraction, feature selection, modeling, matching, and positioning are all included in object recognition [21].

Recently, deep learning, also known as deep machine learning or deep structured learning-based techniques, has seen significant success in digital image processing for object recognition and categorization. Consequently, they are rapidly becoming a focal point of interest among computer vision scientists. There has been a significant rise in the use of digital imaging for tracking marine environments like seagrass beds. As a result, automatic detection and classification based on deep neural networks have become more important tools.

Deep learning's ability to process large amounts of data has the potential to provide solutions to a number of issues pertaining to the marine industry, including marine disaster prevention and mitigation, ecological environmental protection, emergency rescue, and underwater target detection, tracking, and recognition, to mention few of. There are a number of factors that could explain deep learning's comeback, including those listed below:

    - The introduction of large-scale annotated training data, such as those provided by ImageNet [22], to display fully its very vast learning capability;
    - Accelerated development of high-performance parallel computing systems, such as GPU clusters; and
    - Substantial progress made in the development of various network architectures and instructional methods.

The primary contributions of this paper are as follows:

1)   A detailed discussion of the most widely used methods and deep network architectures for the analysis of underwater targets
2)   Large collections of underwater images and video recordings being compiled and studied extensively
3)   A full review and comparison of experiments with different deep learning methods for the detection and recognition of marine objects
4)   Deep learning techniques being used to discuss in depth future trends and possible challenges in recognizing marine objects.

The remainder of this paper is organized as follows: Section 2 presents a Review of Traditional Object Detection Methods that have been used. In Section 3, typical deep learning methods together with comprehensive comparisons are systematically presented. Popular datasets are revisited in Section 4. Previous research methods are discussed in Section 5 and the conclusions are drawn and presented in Section 6.

## 2. REVIEW OF TRADITIONAL OBJECT DETECTION METHODS

The Viola-Jones object detection framework was proposed in 2001 [23, 24]. This framework for face detection is based on the AdaBoost algorithm [25] and uses Haar-like wavelet characteristics and integral graph technology. The combination of Haar and AdaBoost had not hitherto been used in a detection approach. Moreover, it is the first detection framework to operate in real-time. The Viola-Jones detector has been widely used as a foundation for face identification algorithms [26, 27] prior to the development of deep learning technology.

The histogram is computed using the gradient instead of the color value in Histogram of Oriented Gradient (HOG) [3]. The feature is built by computing the local gradient direction histogram of the image. Image recognition applications have made extensive use of HOG features in conjunction with SVM classifiers, particularly for the purpose of pedestrian identification [3]. The invariant histograms of oriented gradients (Ri HOG) [37] use cells of an annular spatial binning type and the radial gradient transform (RGT) to produce gradient binning invariance for feature descriptors, and this is only one example of many related studies. The detection concepts of enhanced HOG, support vector machine classifier, and sliding window are all incorporated into the DPM [29] algorithm, which uses a multicomponent approach to solve the target's multiview problem. In order to address the issue of target deformation, it uses a component model technique with a graphical representation of the target. DPM is a detection method that relies on individual components and has high robustness against target deformation. DPM is the backbone of several deep learning-based algorithms for tasks such as classification, segmentation, posture estimation, etc. [30, 31].

Machine learning-based object detection techniques still have advantages in certain use cases. Data from images were chunked and encoded as vectors in [32]. Sub-features are taken from the color and texture of the images and are then added together to form a feature vector. The use of the Random Forest technique resulted in a classification accuracy of 99.62 percent. By using a 1 master + 4 workers clustering design in Apache Spark, the execution time of each method was accelerated on average by a factor of 3.40.

## 3. DEEP LEARNING-BASED OBJECT DETECTION

We will now investigate various popular state-of-the-art CNN architectures. The convolution layer, the sub-sampling layer, the dense layers, and the soft-max layer form the backbone of the majority of deep convolutional neural networks. The architectures typically consist of stacks of multiple convolutional layers and max-pooling layers followed by fully linked and SoftMax layers at the end. LeNet [33], AlexNet [34], VGG Net [35], NiN [36], and all convolutional (also Conv) [37] are all instances of such models. Other potentially more effective advanced architectures have also been proposed. These include GoogLeNet with Inception units [38, 39], Residual Networks [40], DenseNet [41] and FractalNet [42]. Most of the fundamental building blocks (convolution and pooling) are shared by these many designs. However, newer deep learning architectures have been found to have some topological 'quirks' of their own. In terms of state-of-the-art performance on various benchmarks for object identification tasks, the DCNN designs, namely AlexNet [34], VGG [35], GoogLeNet [38, 39], Dense CNN [41], and FractalNet [42], are widely considered to be the most popular architectures. Some of these architectures (such as GoogLeNet and ResNet) are tailored specifically for processing massive amounts of data, while others (such as the VGG network) are more general in nature. DenseNet [41] is one of the architectures that have a high density of connections. Alternatively, for ResNet, one might try the more flexible Fractal Network.

## 4. DATASETS

Due to the fact that underwater image processing is a relatively new field of study, only a small number of datasets are available for use in underwater computer vision [43]. The following are some of the most important reasons for the small number:

1) Due to a late start in the field, sufficient attention has not been devoted to the relevant underwater image datasets.
2) Although academic researchers have recently begun to recognize the value of an underwater image collection, creating such a dataset is laborious and time-consuming due to the unique challenges presented by the ocean environment.
3) The underwater world is incredibly diverse, making manual collection and classification of ground truths for a wide range of underwater images difficult.

Table 1. Review of some existing databases that can be made available to the general public for underwater object detection.

| Database Name | Introduction |
| --- | --- |
| Underwater Image Enhancement Benchmark (UIEB) [44] | There are 950 genuine underwater images in the UIEB, of which 890 have associated references and 60 do not.<br>The academic goal is to improve underwater images for academic purposes. |
| Marine Underwater Environment Database (MUED) [43] | 430 various classes of interesting objects are represented in MUED's 8,600 underwater images, which vary in stance, position, illumination, turbidity of the water, and more.<br>The academic goal is saliency detection and object recognition in underwater images |
| Real-time Underwater Image Enhancement (RUIE) Dataset [45] | Over 4,000 underwater real images are included in RUIE's Underwater Image Quality Sub-aggregate, Underwater Color Cast Sub-aggregate, and Underwater higher-level task-driven Sub-aggregate.<br>The academic goal has focused on improving underwater images and finding objects in them. |
| The TrashCan dataset [46] | This dataset includes observations of trash, remotely operated vehicles (ROVs), and a diverse range of marine life, all cataloged in a database of annotated images (7,212 images as of this publishing). Instance segmentation annotations are used to label which pixels in the image correspond to which objects in this dataset. collected from a variety of sources. |
| UOT32 (Underwater Object Tracking) Dataset [47] | The benchmark dataset for underwater object tracking has 32 videos with a total of 24,241 annotated frames and an average duration of 29.15 seconds and frame count of 757.53. sequences for objects of interest. |

| SUIM Dataset [48] | This is the first comprehensive dataset for underwater image semantic segmentation (SUIM). Fish (vertebrates), reefs (invertebrates), aquatic plants, wrecks/ruins, human divers, robots, and the seafloor are only a few of the eight object categories covered by more than 1,500 images with pixel annotations. Participants in oceanographic expeditions and human-robot cooperation studies capture and meticulously annotate the images. |
|---|---|
| SeabedObjects-KLSG [49] | A real side-scan sonar image dataset called SeabedObjects-KLSG can be used to identify wrecks, drowning victims, airplanes, mines, and the seafloor. This was done in an effort expeditiously to promote underwater object classification in side-scan sonar images, especially civilian object classification. |
| Fish4K [50] | The resource is referred to as a resource since it comprises sample images of 23 different species. These images are mainly free of noise; however, most are out of focus. |
| Kyutech-10K [51] | This is the first dataset of deep-sea marine organisms provided by the Japan Agency for Marine-Earth Science and Technology (JAMES). |

Figure 1 shows a subset of the 890 identical pairs of original underwater images and reference images that comprise the Underwater Image Enhancement Benchmark (UIEB), and these underwater images are collected from Google, YouTube, related papers and paper researcher self-captured videos [44].

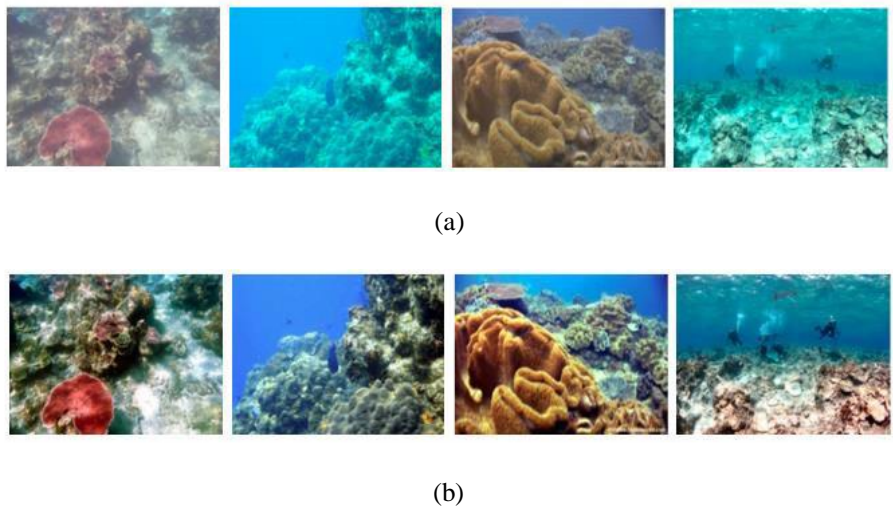

(a)

(b)

Figure 1. Examples from UIEB with subclasses: (a) original underwater images, (b) corresponding reference images.

Some examples of underwater images from MUED [43] with high turbidity, uneven illumination, monotonous hues, and intricate underwater-background are shown in Figure 2. These issues have a significant impact on the reliability and availability of underwater images in real-world applications.

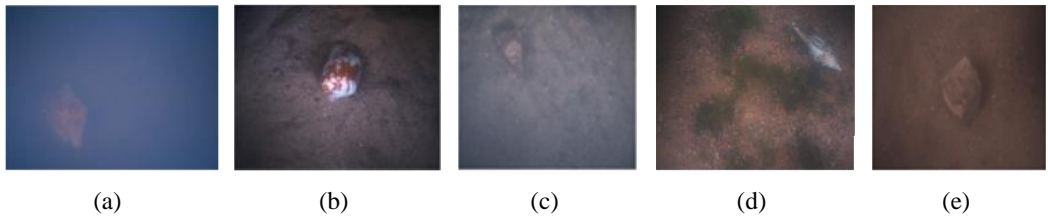

Figure 2. Some examples of detrimental elements present in the marine environment that can affect the use of underwater vision. (a) Water with high turbidity, (b) Uneven illumination, (c) Low contrast, (d) Complicated underwater-background, and (e) Monotonous color

Images captured using an underwater optical imaging and capturing device as part of the Real-time Underwater Image Enhancement (RUIE) Dataset are shown in Figure 3. The Underwater Image Quality Subclass, Underwater Color Cast Subclass, and Underwater higher-level task-driven Subclass are the three subclasses of underwater images that comprise RUIE. In order to gather image examples for the RUIE benchmark, they put up a multi-view underwater image capture system with twenty-two water-proof video cameras.
.

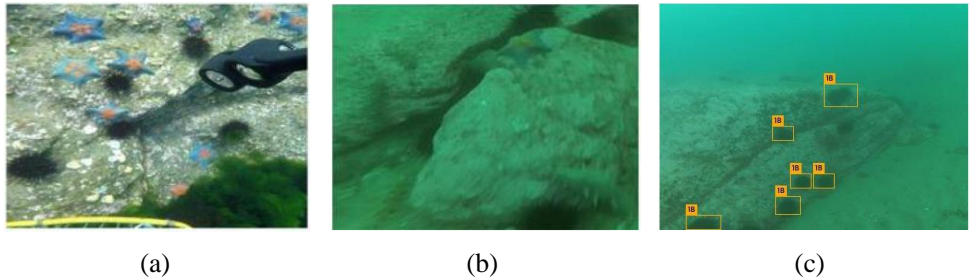

Figure 3. Some images from the RUIE dataset with a triple of subclasses of underwater images: (a) Underwater Image Quality Sub-aggregate, (b) Underwater Color Cast Sub-aggregate, (c) Underwater higher-level task-driven Sub-aggregate.

Figure 4 illustrates a sampling of the results of object detection and instance segmentation models trained on both versions of the datasets [46]. The outcomes encompass an extensive range of object sizes and situations.

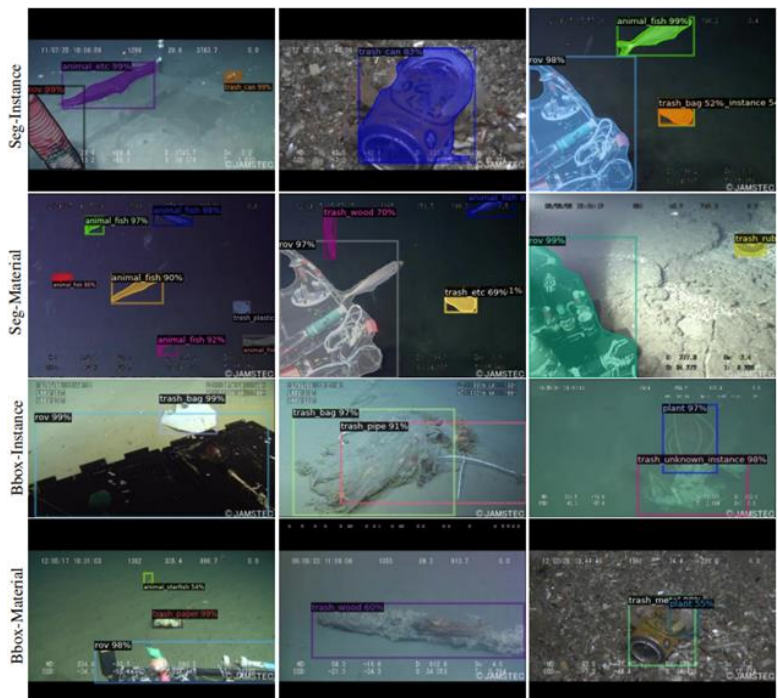

Figure 4. Sampled results for object detection and image segmentation
for both versions of the TrashCan dataset.

The first large-scale, diverse, underwater benchmarking dataset (UOT100) was created with over 74,000 annotated frames spread across 104 video sequences. Both synthetic and natural underwater imagery have similarly distributed aberrations in the dataset as a whole, many different YouTube channels and other internet video platforms contributed to the dataset, as did preposted and manually annotated ground truth bounding box. Figure 5 shows a visual summary of the distortions as categories that represent the color of the water, such as blue, green, and yellow.

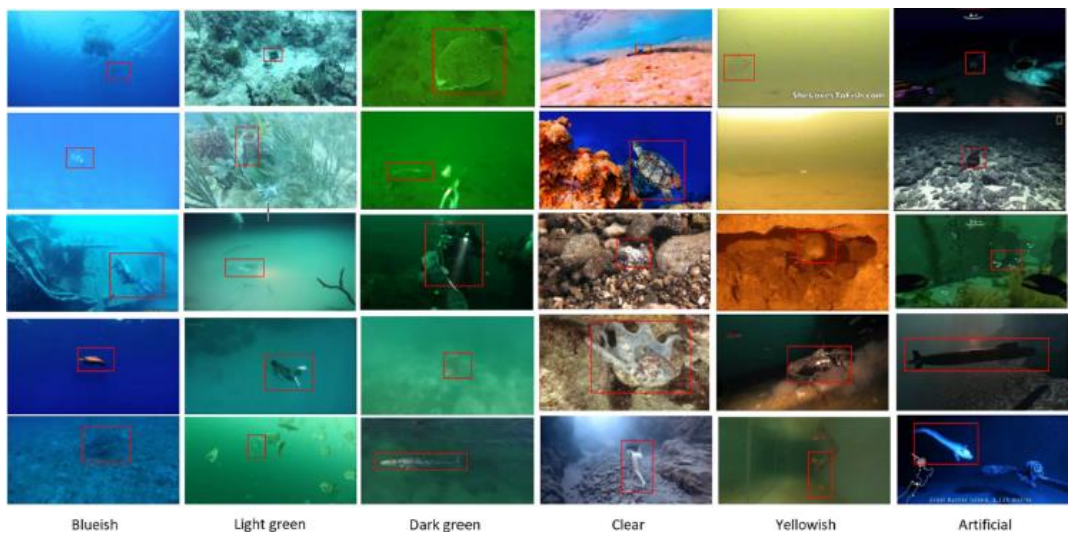

Figure 5. Sample tracking data from our UOT100 dataset showing various types of distortions. The red bounding boxes denote the object of interest and the text below each column indicates the category of the visual data

In total there are 1,525 RGB images in the SUIM dataset that may be used for either training or validation, and an additional 110 test images can be used as a benchmark for assessing the performance of semantic segmentation models. There is a wide range of spatial resolutions present in the photos, including $256 \times 256$, $640 \times 480$, $1280 \times 720$ and $1906 \times 1080$. Seven human volunteers labeled every pixel of the SUIM dataset. An example or two can be seen in Figure 4.6.

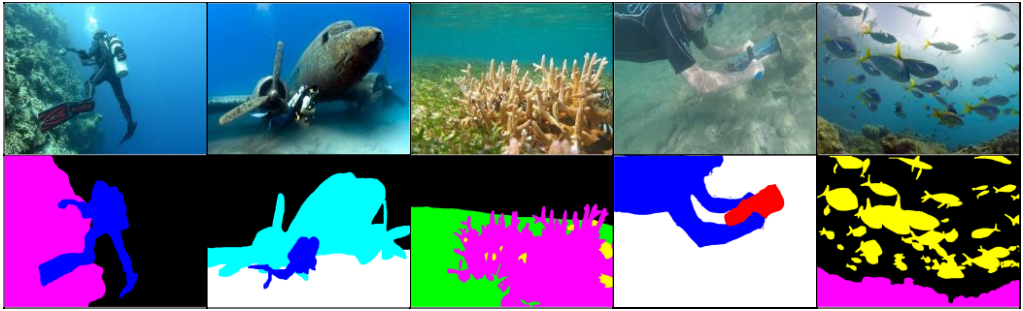

Figure 6. A few sample images and corresponding pixel-annotations are shown on the top and bottom rows, respectively

There are currently 385 wreck images, 36 drowning victim images, 62 aircraft images, 129 mine images, and 578 seafloor images in the dataset known as SeabedObjects-KLSG. All of the images were taken directly from the raw data of the large sides can sonar images. Figure 7 shows some data from the SeabedObjects-KLSG dataset.

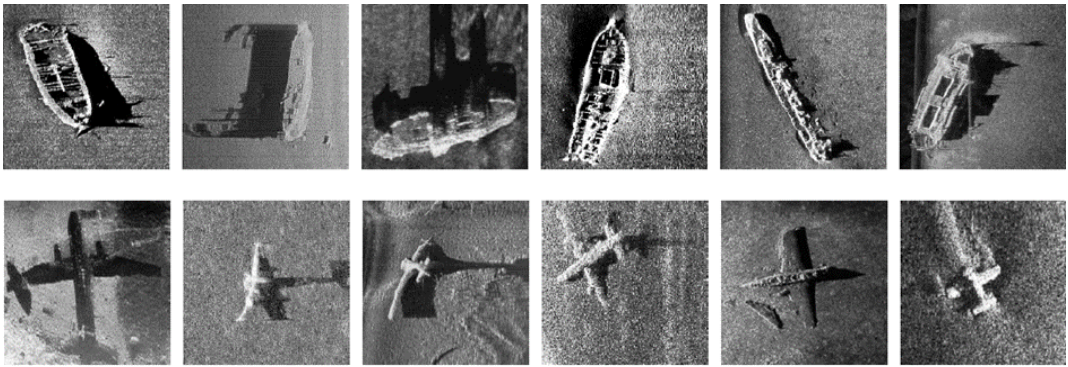

Figure 7. Samples from the SeabedObjects KLSG dataset.

Research on marine ecosystems is aided by the Fish4Knowledge dataset, which was released by the Taiwan Ocean Research Institute and numerous other partner institutes. Figure 8 depicts a handful of images from the dataset consisting of 27,370 tagged underwater images of 23 distinct fish species acquired over the course of two years by 10 underwater cameras in Taiwanese inland lakes.

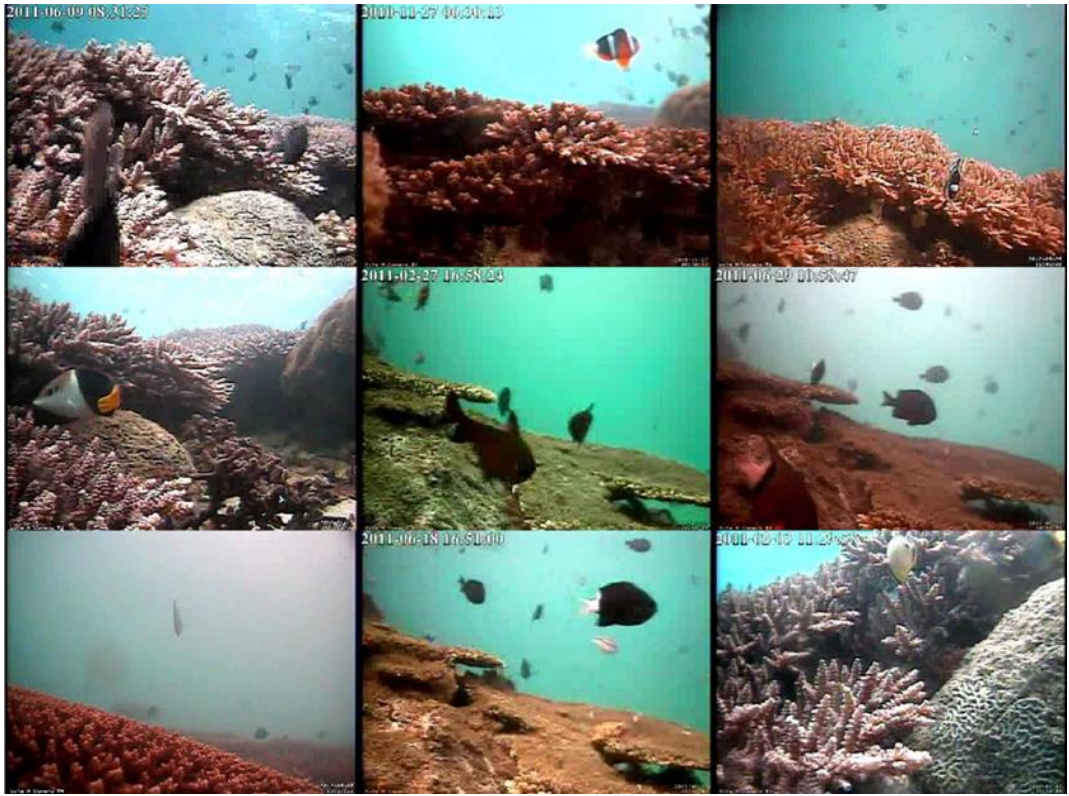

Figure 8. Examples of underwater images on a Taiwan reef with different background variability.

Kyutech10K has 10,728 images and 1,489 videos over seven different categories (shrimp, squid, crab, shark, sea urchin, manganese and sand). Every still image and video clip will always be displayed at a maximum resolution of 480 × 640 pixels. In Figure 9, we provide a sample of images for each group.

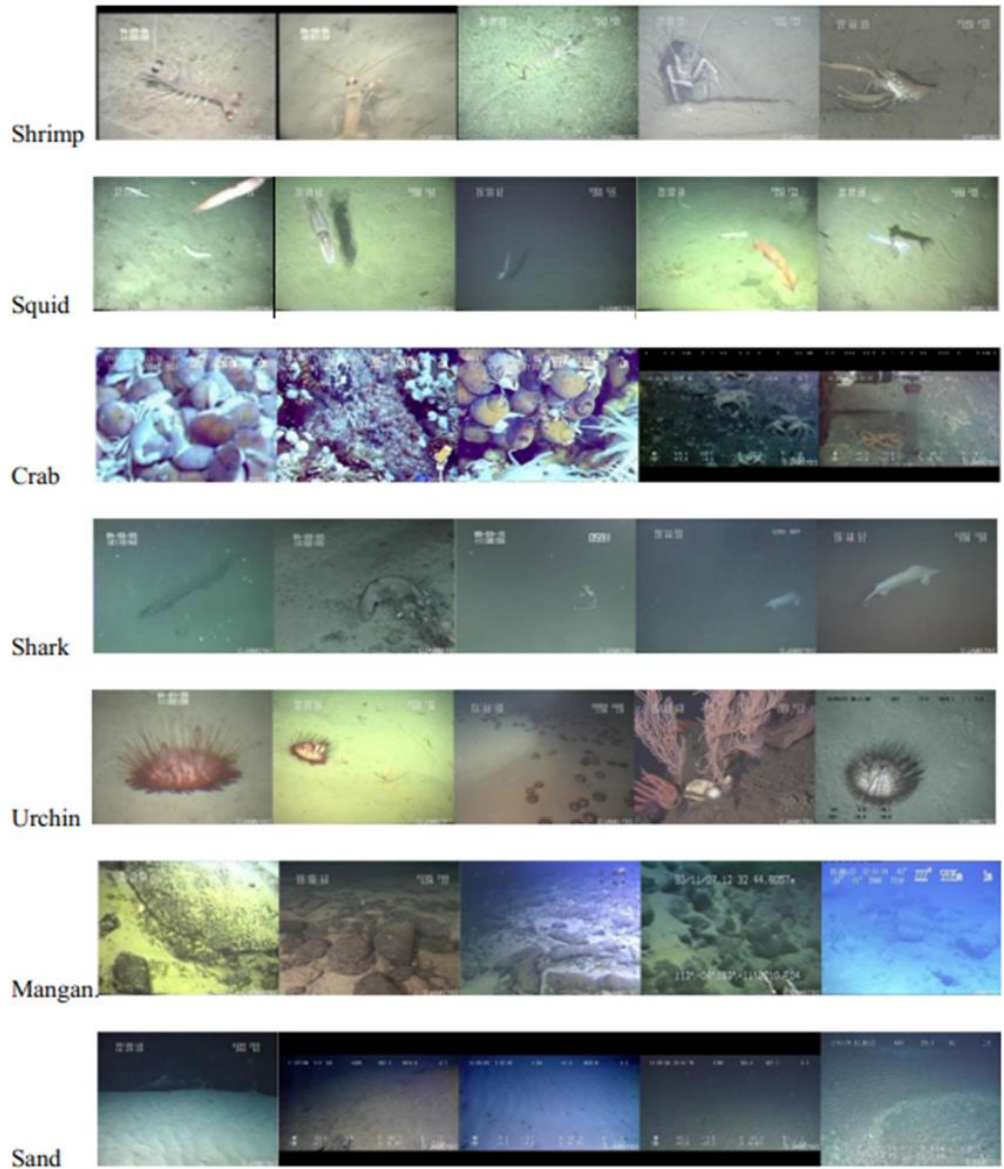

Figure 9. The Kyutech10K dataset.

## 5. PREVIOUS RESEARCH METHODS

It has been shown that a deep Convolutional Neural Network, such as the one proposed by Nicole Seese et al. [52], performs admirably in a dynamic setting, hence these researchers proposed an Adaptive Foreground Extraction Method using a deep Convolution Neural Network for classification. Because of its emphasis on lighting uncertainty, background motion and non-static imaging platforms, it performs well in practical settings. A Gaussian Mixture Model is employed in dynamic settings, while a Kalman filter is reserved for less complex circumstances. Therefore, the method's efficiency and speed are likely to deteriorate.

The paper by Xiu Li1 Min Shang et al. [53] uses a fast R CNN approach designed specifically for the detection of fish. The approach returns values with higher mean average precision and is faster than R CNN (map). In total, the study contributed to the creation of a brand-new, massive dataset consisting of 242,722 images over 12 distinct classes. Time-consuming selective search is used to collect the input of 2,000 regions of interest (ROI) for the network. Despite its rapidity, this operation is not real-time.

With regard to hybrid features, the deep learning method utilizing VGGNet presented by A. Mahmood et al. [54] proposes an extraction strategy based on the Spatial Pyramid Pooling (SPP) approach, in which a pre-trained VGGNet is used to improve categorization by combining deep features from the VGGNet with texton and color-based characteristics. The CNN is then trained using the MLC dataset.

More accurate detection of zooplankton with the Convolutional Neural Network-based ZooplanktonNet model is presented by Jialun Dai et al. [55]. In order to reduce overfitting, it leverages augmentation of existing data to make the classification process more accurate. CNN is a more efficient image classification system since it does not rely on training data or previous samples. Despite there being an insufficient number of zooplankton images to train deep neural networks, this study appeared to work well with less knowledge.

In order to achieve fine-grained classification using a CNN, Hansang Lee et al. [56] combine transfer learning with a pre-trained CNN. A combination of data augmentation methods, including transfer learning, was employed to correct the issue of class imbalance. It is applicable and efficient to produce a satisfying outcome, and it is particularly useful for large-scale class imbalance datasets.

Sebastien Villon et al. [57] proposed a combination of Convolutional and Deep Learning techniques, a Neural Network, and HOG+SVM to detect submerged objects. This combination is able to identify coral reef fish from video stills taken underwater. The study titled "A Comparative Study of Robust Underwater Object Detection with Autonomous Underwater Vehicle" ICCA 2020, Dhaka, Bangladesh found that deep learning yields better detection accuracy than conventional approaches. With the use of image contours, HOG can uncover intricate situations that are otherwise obscured, such as those hiding in coral reefs.

Convolutional neural networks (CNNs) with a global average pooling (GAP) layer before each fully connected layer to generate a class activation map were proposed by Gebhardt et al. in [58]. To locate MLOs in sidescan sonar images, the researchers in [58] used a DNN. The authors examined how several factors, including DNN depth, memory, calculation, and training data distribution, affected detection performance. Furthermore, they used visualization methods to make the model's behavior more understandable to end users. Complex DNN models produce higher accuracy (98%) than simple DNN models (93%) and perform better (78%) than SVM models. The most complex DNN models improved performance by 1 percent but required 17 times as many trainable parameters to do so. The described method uses less computing power than DNNs designed for multi-class classification workloads. For this reason, it can be used by unmanned marine vehicles.

In order to perform semantic segmentation, the SegNet [59] uses a fully convolutional encoder-decoder architecture. All thirteen convolutional layers used by the VGG16 image classifier are replicated topologically in its encoder network. The SegNet's decoder network architecture allows for far less memory to be used, which is its primary advantage over alternative segmentation systems. Since the SegNet is a traditional CNN-based image segmentation architecture, we deemed it to be a good candidate for evaluation.

Table 2. Review of existing databases for underwater object detection that
can be made accessible to the public.

| Method | Advantages | Disadvantages |
|---|---|---|
| Adaptive Foreground Extraction Method [52] | CNN for classification works well in dynamic environments, focuses on uncertain illumination factors and non-static environments, and works well in dynamic environments. | Use of the Gaussian Mixture Model and the Kalman Filter, both of which diminish speed and efficiency, should be relegated to more complicated and dynamic circumstances. |
| R-CNN [60] | Utilizes a filtered search in order to generate regions. Approximately two thousand regions are retrieved from each image. | Because each region is handed over to the CNN model on an individual basis, a significant amount of processing time is consumed. In addition, it uses three distinct networks to make predictions. |
| Fast R-CNN [53] | Faster than R-CNN, the dataset for recreation of fish. The CNN model only has to be trained once with each image before extracting feature maps. Predictions are generated via a selective search on these feature maps. It utilizes all three models used by R-CNN. | The use of 2,000 regions of interest as input necessitates a significant amount of startup time and is therefore inapplicable to real-life scenarios |
| Faster RCNN [60] | Selective search has been replaced in this model by the use of a technique called Region Proposal Network (RPN). In comparison to the other versions listed above, RPN increases the speed of the model significantly. | -To successfully extract all items from a single image, the method requires multiple iterations. -Due to the sequential nature of these algorithms, the success of subsequent stages of the network is contingent on the results of previous systems. |
| VGGNet [54] | Features are hybrid and deep features are used for pre-training. | Utilization of the MLC Dataset, which is inappropriate for use in image classification. |
| ZooplanktonNet [55] | A high accuracy rate, the use of data augmentation to reduce the amount of data overfitting, and reduced preprocessing are all features of this model. | The absence of images of plankton, which is necessary for a deep neural network, which requires massive datasets. |
| CNN+Transfer Learning [56] | Pre-trained CNN, overcoming the class imbalance problem, use of numerous data augmentation approaches. | Optimal for massive data-intensive tasks, not at all for more modest endeavors. |

| HOG+SVM [57] | Used for locating submerged items that may otherwise go undetected. | More time-consuming and less effective than deep learning approaches in terms of both detection and efficiency. |
|---|---|---|
| Different structures of convolutional neural networks (CNNs) [58] | - High accuracy (93%) <br> - Can be used with self-driving underwater vehicles | When compared to DNNs designed for multi-class classification applications, the computational requirements of the proposed method are lower. |
| SegNet [59] | Decoder network's capacity severely reduces RAM consumption. | The precision of feature extraction is linearly proportional to the complexity of the model. |

## 6. CONCLUSIONS

Because of its promise, deep learning has already altered many facets of public life. Generic object detection has been quite successful thanks to the availability of large amounts of data and powerful computers. The field of marine engineering has focused much attention in recent years to methods of detecting objects submerged in the ocean using deep learning. This can be used for a variety of marine pursuits. Based on the current state of the art in underwater object identification research, this study provides a thorough categorization and analysis of relevant publications. Well-known reference datasets have been covered. A comparison is made between various deep learning methods and more traditional methods. The ideal approach for underwater item detection seems to be the Convolutional Neural Networks (CNN), which are generally regarded for computer vision models and classification in complicated situations. The goal of this article is to provide readers with a thorough understanding of the current state of underwater object detection in the hope that it will help them in their own research endeavors.

## REFERENCES

[1] Wu, H., Q. Liu, and X. Liu, A review on deep learning approaches to image classification and object segmentation. TSP, 2018. 1(1): p. 1-5.

[2] Lowe, D.G.J.I.j.o.c.v., Distinctive image features from scale-invariant keypoints. 2004. 60(2): p. 91-110.

[3] Dalal, N. and B. Triggs. Histograms of oriented gradients for human detection. in 2005 IEEE computer society conference on computer vision and pattern recognition (CVPR'05). 2005. Ieee.

[4] Lienhart, R. and J. Maydt. An extended set of haar-like features for rapid object detection. in Proceedings. international conference on image processing. 2002. IEEE.

[5] Cortes, C. and V.J.M.l. Vapnik, Support vector machine. 1995. 20(3): p. 273-297.

[6] Freund, Y., R.E.J.J.o.c. Schapire, and s. sciences, A decision-theoretic generalization of on-line learning and an application to boosting. 1997. 55(1): p. 119-139.

[7] Felzenszwalb, P.F., et al., Object detection with discriminatively trained part-based models. 2010. 32(9): p. 1627-1645.

[8] Yang, H., et al., Research on underwater object recognition based on YOLOv3. Microsystem Technologies, 2021. 27(4): p. 1837-1844.

[9] LeCun, Y., et al., Backpropagation applied to handwritten zip code recognition. Neural computation, 1989. 1(4): p. 541-551.

[10] LeCun, Y., et al., Handwritten digit recognition with a back-propagation network. Advances in neural information processing systems, 1989. 2.

[11] Anagnostopoulos, C.-N.E., et al., License plate recognition from still images and video sequences: A survey. IEEE Transactions on intelligent transportation systems, 2008. 9(3): p. 377-391.

[12] El Ayadi, M., M.S. Kamel, and F. Karray, Survey on speech emotion recognition: Features, classification schemes, and databases. Pattern recognition, 2011. 44(3): p. 572-587.

[13] Borkar, A., M. Hayes, and M.T. Smith, A novel lane detection system with efficient ground truth generation. IEEE Transactions on Intelligent Transportation Systems, 2011. 13(1): p. 365-374.

[14] Liu, W., et al. Sphereface: Deep hypersphere embedding for face recognition. in Proceedings of the IEEE conference on computer vision and pattern recognition. 2017.

[15] Yang, X., P. Molchanov, and J. Kautz. Making convolutional networks recurrent for visual sequence learning. in Proceedings of the IEEE Conference on Computer Vision and Pattern Recognition. 2018.

[16] Zabidi, M.M., et al. Embedded vision systems for ship recognition. in TENCON 2009-2009 IEEE region 10 conference. 2009. IEEE.

[17] Jin, L. and H. Liang. Deep learning for underwater image recognition in small sample size situations. in OCEANS 2017-Aberdeen. 2017. IEEE.

[18] Meng, L., T. Hirayama, and S.J.I.A. Oyanagi, Underwater-drone with panoramic camera for automatic fish recognition based on deep learning. 2018. 6: p. 17880-17886.

[19] Yuh, J., G. Marani, and D.R.J.I.s.r. Blidberg, Applications of marine robotic vehicles. 2011. 4(4): p. 221-231.

[20] Liu, Z., et al., Unmanned surface vehicles: An overview of developments and challenges. 2016. 41: p. 71-93.

[21] Yang, H., et al., Research on underwater object recognition based on YOLOv3. 2021. 27(4): p. 1837-1844.

[22] Deng, J., et al. Imagenet: A large-scale hierarchical image database. in 2009 IEEE conference on computer vision and pattern recognition. 2009. Ieee.

[23] Viola, P. and M. Jones. Rapid object detection using a boosted cascade of simple features. in Proceedings of the 2001 IEEE computer society conference on computer vision and pattern recognition. CVPR 2001. 2001. Ieee.

[24] Viola, P. and M.J.J.I.j.o.c.v. Jones, Robust real-time face detection. 2004. 57(2): p. 137-154.

[25] Rätsch, G., T. Onoda, and K.-R.J.M.l. Müller, Soft margins for AdaBoost. 2001. 42(3): p. 287-320.

[26] Yang, B., et al. Aggregate channel features for multi-view face detection. in IEEE international joint conference on biometrics. 2014. IEEE.

[27] Cerf, M., et al., Predicting human gaze using low-level saliency combined with face detection. 2007. 20.

[28] Luo, Z., et al. Rotation-invariant histograms of oriented gradients for local patch robust representation. in 2015 Asia-Pacific Signal and Information Processing Association Annual Summit and Conference (APSIPA). 2015. IEEE.

[29] Felzenszwalb, P., D. McAllester, and D. Ramanan. A discriminatively trained, multiscale, deformable part model. in 2008 IEEE conference on computer vision and pattern recognition. 2008. Ieee.

[30] Liu, W., et al. Ssd: Single shot multibox detector. in European conference on computer vision. 2016. Springer.

[31] Newell, A., K. Yang, and J. Deng. Stacked hourglass networks for human pose estimation. in European conference on computer vision. 2016. Springer.

[32] DOLAPCI, B., C.J.J.o.I.S.T. ÖZCAN, and Applications, Automatic ship detection and classification using machine learning from remote sensing images on Apache Spark. 2021. 4(2): p. 94-102.

[33] LeCun, Y., et al., Gradient-based learning applied to document recognition. 1998. 86(11): p. 2278-2324.

[34] Krizhevsky, A., I. Sutskever, and G.E.J.A.i.n.i.p.s. Hinton, Imagenet classification with deep convolutional neural networks. 2012. 25.

[35] Simonyan, K. and A.J.a.p.a. Zisserman, Very deep convolutional networks for large-scale image recognition. 2014.

[36] Lin, M., Q. Chen, and S.J.a.p.a. Yan, Network in network. 2013.

[37] Springenberg, J.T., et al., Striving for simplicity: The all convolutional net. 2014.

[38] Szegedy, C., et al. Going deeper with convolutions. in Proceedings of the IEEE conference on computer vision and pattern recognition. 2015.

[39] Szegedy, C., et al. Inception-v4, inception-resnet and the impact of residual connections on learning. in Thirty-first AAAI conference on artificial intelligence. 2017.

[40] He, K., et al. Deep residual learning for image recognition. in Proceedings of the IEEE conference on computer vision and pattern recognition. 2016.

[41] Huang, G., et al. Densely connected convolutional networks. in Proceedings of the IEEE conference on computer vision and pattern recognition. 2017.

[42] Larsson, G., M. Maire, and G.J.a.p.a. Shakhnarovich, Fractalnet: Ultra-deep neural networks without residuals. 2016.

[43] Jian, M., et al., The extended marine underwater environment database and baseline evaluations. 2019. 80: p. 425-437.

[44] Li, C., et al., An underwater image enhancement benchmark dataset and beyond. 2019. 29: p. 4376-4389.

[45] Liu, R., et al., Real-world underwater enhancement: Challenges, benchmarks, and solutions under natural light. 2020. 30(12): p. 4861-4875.

[46] Hong, J., M. Fulton, and J.J.a.p.a. Sattar, Trashcan: A semantically-segmented dataset towards visual detection of marine debris. 2020.

[47] Kezebou, L., et al. Underwater object tracking benchmark and dataset. in 2019 IEEE International Symposium on Technologies for Homeland Security (HST). 2019. IEEE.

[48] Islam, M.J., et al. Semantic segmentation of underwater imagery: Dataset and benchmark. in 2020 IEEE/RSJ International Conference on Intelligent Robots and Systems (IROS). 2020. IEEE.

[49] Huo, G., Z. Wu, and J.J.I.a. Li, Underwater object classification in sidescan sonar images using deep transfer learning and semisynthetic training data. 2020. 8: p. 47407-47418.

[50] Lines, J., et al., An automatic image-based system for estimating the mass of free-swimming fish. 2001. 31(2): p. 151-168.

[51] Lu, H., et al., FDCNet: filtering deep convolutional network for marine organism classification. 2018. 77(17): p. 21847-21860.

[52] Seese, N., et al. Adaptive foreground extraction for deep fish classification. in 2016 ICPR 2nd Workshop on Computer Vision for Analysis of Underwater Imagery (CVAUI). 2016. IEEE.

[53] Li, X., et al. Fast accurate fish detection and recognition of underwater images with fast r-cnn. in OCEANS 2015-MTS/IEEE Washington. 2015. IEEE.

[54] Mahmood, A., et al. Coral classification with hybrid feature representations. in 2016 IEEE International Conference on Image Processing (ICIP). 2016. IEEE.

[55] Dai, J., et al. ZooplanktoNet: Deep convolutional network for zooplankton classification. in OCEANS 2016-Shanghai. 2016. IEEE.

[56] Lee, H., M. Park, and J. Kim. Plankton classification on imbalanced large scale database via convolutional neural networks with transfer learning. in 2016 IEEE international conference on image processing (ICIP). 2016. IEEE.

[57] Villon, S., et al. Coral reef fish detection and recognition in underwater videos by supervised machine learning: Comparison between Deep Learning and HOG+ SVM methods. in International Conference on Advanced Concepts for Intelligent Vision Systems. 2016. Springer.

[58] Gebhardt, D., et al. Hunting for naval mines with deep neural networks. in OCEANS 2017-Anchorage. 2017. IEEE.

[59] Badrinarayanan, V., et al., Segnet: A deep convolutional encoder-decoder architecture for image segmentation. 2017. 39(12): p. 2481-2495.

[60] Fayaz, S., et al., Underwater object detection: architectures and algorithms–a comprehensive review. 2022: p. 1-46.